\documentclass[10pt,twocolumn,letterpaper]{article}

\usepackage{cvpr}
\usepackage{times}
\usepackage{epsfig}
\usepackage{graphicx}
\usepackage{amsmath}
\usepackage{amssymb}

\DeclareMathOperator*{\argmin}{\arg\!\min}
\usepackage[pagebackref=true,breaklinks=true,letterpaper=true,colorlinks,bookmarks=false]{hyperref}

\cvprfinalcopy % *** Uncomment this line for the final submission

\def\cvprPaperID{0035} % *** Enter the CVPR Paper ID here

\ifcvprfinal\fi
\begin{document}

%%%%%%%%% TITLE
\title{Domain Adaptation with Soft-margin multiple feature-kernel learning beats Deep Learning for surveillance face recognition}

\author{Samik Banerjee\\
IIT Madras\\
Chennai, 600036, India\\
{\tt\small samik@cse.iitm.ac.in}
% For a paper whose authors are all at the same institution,
% omit the following lines up until the closing ``}''.
% Additional authors and addresses can be added with ``\and'',
% just like the second author.
% To save space, use either the email address or home page, not both
\and
Sukhendu Das\\
IIT Madras\\
Chennai, 600036, India\\
{\tt\small sdas@iitm.ac.in}
}

\maketitle
%\thispagestyle{empty}

%%%%%%%%% ABSTRACT
\begin{abstract}
Face recognition (FR) is the most preferred mode for biometric-based surveillance, due to its passive nature of detecting subjects, amongst all different types of biometric traits. FR under surveillance scenario does not give satisfactory performance due to low contrast, noise and poor illumination conditions on probes, as compared to the training samples. A state-of-the-art technology, Deep Learning, even fails to perform well in these scenarios. We propose a novel soft-margin based learning method for multiple feature-kernel combinations, followed by feature transformed using Domain Adaptation, which outperforms many recent state-of-the-art techniques, when tested using three real-world surveillance face datasets.
\end{abstract}

%%%%%%%%% BODY TEXT
\section{Introduction}
\label{sec:intro}
Face recognition (FR) has gained importance in security for surveillance, which has recently attracted the attention of vision researchers. In surveillance scenarios, the problem of face recognition is difficult as the image frames captured by CCTV cameras often suffer from low illumination and distance-based attenuation factors, thus capturing poor quality images. The low quality face images coupled with low resolution produce unsatisfactory performance (accuracy) of recognition. There has been several attempts of FR to solve this problem in the recent past. Traditional FR techniques fail to cater to the changes in the resolution, contrast and illumination between the gallery and the probe samples.

The proposed approach consists of three stages: (i) Pre-processing and feature-extraction, (ii) Soft-margin learning for multiple feature-kernel combinations (SML-MFKC), and (iii) Domain Adaptation (DA). In the rest of the paper, section \ref{sec:rw} details the recent advances in the field of the face detection and recognition under surveillance scenario, multiple kernel learning based classification and domain adaptation. Section \ref{sec:prop} briefly outlines the proposed techniques, while the section \ref{sec:data} introduces the three real world surveillance datasets used for the evaluation of the proposed techniques. Finally, section \ref{sec:exp} gives the quantitative results of performance analysis for the proposed technique, on the three datasets, using rank-1 recognition rate, CMC and ROC metrics, and section \ref{sec:conc} concludes the paper. 

\section{Related Works}
\label{sec:rw}

The most widely used face detection algorithm proposed by Viola \etal \cite{viola2004robust}, is based on an efficient classifier build using the ADABOOST learning algorithm, which selects a subset of critical visual features from a very large set of potential features. Our proposed technique includes the face detection technique based on the set of $49$ fiducial landmark points detected by the Chehra \cite{asthanaincremental} face detector.

A large scale implementation of support vector machine (SVM) for large number of kernels, known as sequential minimal optimization (SMO), was proposed by Bach \etal \cite{bach2004multiple}. A multiple kernel learning (MKL) algorithm based on sparse representation-based classification (SRC) proposed in \cite{shrivastava2014multiple}, represents the non-linearities in the high-dimensional feature space based on kernel alignment criteria. Conic combinations of kernel matrices for classification proposed in \cite{lanckriet2004statistical} leads to a convex quadratically constrained quadratic problem (QCQP). Sonnenburg \etal \cite{sonnenburg2006large} generalized the formulation to a larger class of problem.

The work proposed in \cite{kliep} performs domain adaptation based on the calculation of the weights of the instances in the source domain. Yang \etal \cite{yang2007cross} proposed a method to effectively retrain a pre-trained SVM for target domain data, based on the calculated weights of the instances in the source domain. Duan \etal \cite{duan2012domain} proposed a domain adaptive machine (DAM), which learns a robust decision function for labeling the instances in the target domain, by leveraging a set of base classifiers learned on multiple source domains. Tranfer component analysis (TCA), proposed in \cite{pan2011domain}, minimizes the disparity of distribution by comparing the difference in the means between two domains and preserving the local geometry of the underlying manifold. Subspaces are calculated based on eigen-vectors \cite{fernando2013unsupervised} of two domains, such that the basis vectors of the of the transformed source and target domains are aligned. Wang \etal \cite{wang2011heterogeneous} considered the manifold of each domain and estimated a latent space, where the manifolds of both the domains are similar to each other. A new method of unsupervised DA was proposed by Samanta \etal \cite{samanta2015unsupervised} using the properties of the of the sub-spaces spanning the source and target domains, when projected along a path in the Grassmannian manifold.

Surveillance cameras produce images at very low resolution to cope with the high transmission speed and optimality of the storage data. Besides the degradation incurred due to low resolution, the images captured by surveillance cameras also contain noise and distortions (defocus, blur, low contrast), due to their uncontrolled environment conditions during capture. A matching algorithm based on transformation learning through an iterative majorization algorithm, in the kernel space, was proposed by Biswas \etal \cite{biswas2012multidimensional}, known as multi-dimensional scaling (MDS). Ren \etal \cite{ren2012coupled} proposed the Coupled Kernel Embedding approach, where they map the low and high resolution face images onto different kernel spaces and then transform them to a subspace for recognition. Rudrani \etal in \cite{rudrani2011face} proposed an approach with the combination of partial restoration (using super-resolution) of probe samples and degradation of gallery samples. An outdoor surveillance dataset, FR\_SURV, was also proposed in \cite{rudrani2011face}, for evaluating their approach. A Dynamic Bayesian Network (DBN) based unconstrained face recognition under surveillance scenario has been proposed by An \etal \cite{an2013dynamic} to integrate the information from all the three cameras in the ChokePoint \cite{wong_cvprw_2011} dataset. The work proposed in \cite{banerjee2014face} aims to bridge the gap of resolution and contrast using super-resolution and contrast stretching on the probe samples and degrading the gallery samples. A DA technique based on an eigen-domain transformation was proposed to make the distributions of gallery (as source) features identical to that of probe (as target) samples. 

\section{Proposed Method}
\label{sec:prop}
\begin{figure} [!htbp]
    \centering
    \includegraphics[scale=0.27]{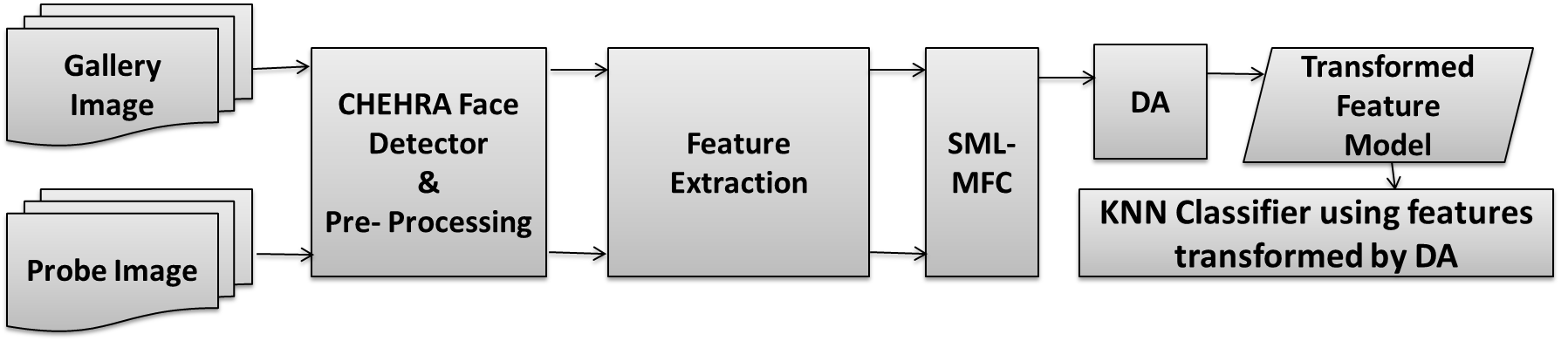} 
    \caption{Overall proposed framework of SML-MKFC with DA.}
    \label{fig:fw}
\end{figure}

The stages of the overall proposed framework for FR, as shown in figure \ref{fig:fw}, is briefly discussed below:

\subsection{Face Detection}
The Face detection stage is based on the 49 fiducial points obtained from the Chehra \cite{asthanaincremental} face detector, to obtain a tightly cropped facial region.

\subsection{Pre-processing}
\begin{figure} [!htbp]
\centering
\includegraphics[scale=0.35]{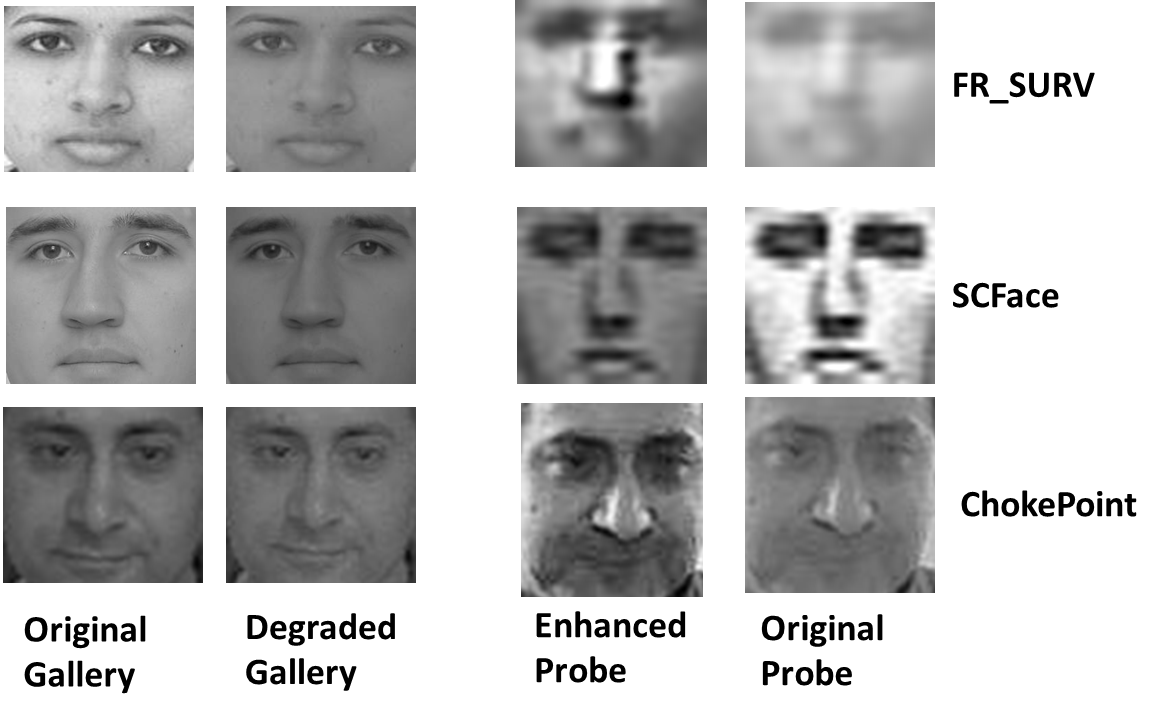}
\caption{Example shots from the three datasets (one sample each), showing the degraded gallery and the enhanced probe images on the cropped faces provided by Chehra \cite{asthanaincremental}.}
\label{fig:gp}
\end{figure}

Pre-processing of both the gallery and probe samples are required  to bridge the gap between the face images obtained from the gallery and probe samples. The stages of pre-processing are the same as described in \cite{banerjee2014face}. The empirical values of the parameters in the pre-processing algorithms (see \cite{banerjee2014face} for details) are: the Gaussian blur kernel, $\sigma$, for degradation of the gallery; and $\gamma$ used for contrast enhancement of the probes, are given in table \ref{tab:param}, for the three datasets used for performance analysis. An example showing the degraded gallery and enhanced probe for each dataset is shown in figure \ref{fig:gp}.

%\vspace{-4mm}
\begin{table}[!htbp]
\centering
\caption{Values of $\sigma$ for gallery degradation and $\gamma$ for the contrast-stretching in probe enhancement. For details see \cite{banerjee2014face}.}
\begin{tabular}{@{}|p{2.5cm}|p{.5cm}|p{.5cm}|} 
  \hline
  % after \\: \hline or \cline{col1-col2} \cline{col3-col4} ...
  \textbf{Datasets} & \textbf{$\sigma$} & \textbf{$\gamma$}\\\hline
  FR\_SURV \cite{rudrani2011face} & 1.75 & 1.75 \\\hline
  SCFace \cite{grgic2011scface} & 1.70 & 1.50 \\\hline
  ChokePoint \cite{wong_cvprw_2011} & 1.20 &  1.25 \\\hline
\end{tabular}
\label{tab:param}
\end{table}

\vspace{-6mm}

\vspace{-5mm}
\subsection{Feature Extraction}
Holistic features are extracted from each of these gallery and probe face samples. Several recent state-of-the art as-well-as traditional features are taken into account viz., Eigenfaces \cite{turk1991eigenfaces}, Fisherfaces \cite{belhumeur1997eigenfaces}, Weberfaces \cite{wang2011illumination}, Local binary pattern (LBP) \cite{ahonen2006face}, Gaborfaces \cite{liu2002gabor}, Bag-of-words (BOW) \cite{filliat2007visual}, Fisher vector encoding on dense-SIFT features (FV-SIFT) \cite{perronnin2010improving} and VLAD encoding on dense-SIFT features (VLAD-SIFT) \cite{arandjelovic2013all}. These form a bank of $8$ sets of feature descriptors.

\subsection{Soft-margin learning for multiple feature-kernel combinations (SML-MFKC)}
The process of selecting the the best performing kernel function and its parameters in a support vector machine (SVM) during training, generally consists of a cross-validation procedure. MKL techniques have been used to cope with this, where instead of selecting a specific kernel function, multiple kernels are learned using a weighted combination along with its corresponding parameters. Our proposed technique takes into consideration both the kernel and feature to be selected from a single framework.  Given a training set ${(x_i, y_i)}, \forall i=1,...,N$, of $N$ instances, each consisting of an image $x_i \in X$ and a class label $y_i \in {1,..., C}$, and given a set of $F$ image features $V_m : X \rightarrow{R^{d_m}}, \forall m = 1,...,F$, where $d_m$ denotes the dimensionality of the $m$-th feature, the problem of learning a classification function $y : X \rightarrow{1,..., C}$ from the features and training set is called the feature combination problem \cite{gehler2009feature}.

Since we associate image features with kernel functions ($k_m$), kernel selection translates naturally into a feature selection problem. The objective of SML-MFKC is to jointly optimize over a linear combination of kernels $k^{*}(x,y) = \sum^{F}_{m=1} \beta_m k_m(x,y)$, $\beta_m \in \mathcal{R}$, $\vec{\beta} \in \mathcal{R}^F$ and the parameters $\vec{\alpha} \in \mathcal{R}^N$ and $b \in \mathcal{R}$ of an SVM. The objective function which determines the optimal combination of the feature and the kernel for the method, is given by the following proposed (novel) soft-margin cost function:
\vspace{-4mm}
\begin{equation}
\label{eq:SML-MFKC}
\begin{split}
\sup \argmin_{\vec{\beta}^q \in \mathcal{R}^F} & \hspace{2mm} \frac{1}{2} \sum_{m=1}^{F} \beta_m^q \vec{\alpha}^T V_m^q \vec{\alpha} \\
& + C  \sum_{i=1}^{N} L(y_i, b +  \sum_{m=1}^{F} \beta_m^q \vec{\alpha}^T V_m^q \vec{\alpha})\\
\text{s.t} &\hspace{2mm}  \sum_{m=1}^{F} \beta_m^q=1, \beta_m^q \ge 0, \forall m, q = 1, ... P.\\
\end{split}
\end{equation}

where, $L(y,t) = max(0,1-y_t)$ denotes the \textit{Hinge Loss}, $P$  denotes the total number kernels used for learning (see table \ref{tab:ker_typ}, for details of the six different kernels used), $V_m^q$ denotes the $m$-th feature for the $q$-th kernel function, and $\beta_m^q \in \mathcal{R}$ denotes the weight coefficient for the $m$-th feature and the $q$-th kernel combination. 

\begin{table}[!htbp]
\caption{Different kernels used in SML-MFKC and their formulae.}
\begin{tabular}{|p{2.5cm}|p{5cm}|}
\hline
\textbf{Types of Kernel function} & \textbf{Formula} \\
\hline
Linear & $k(x,y) = x^T y + c$ \\
\hline
Polynomial & $k(x,y) = (\alpha x^T y + c) ^d$ \\
\hline
Gaussian & $k(x, y) = exp \left(-\frac{\| x-y \| ^2}{2\sigma^2} \right)$ \\
\hline
RBF & $k(x, y) = exp \left(-\frac{\| x-y \|}{2\sigma^2} \right)$ \\
\hline
Chi-square & $k(x,y) = 1 - \Sigma_{i=1}^n \frac{(x_i-y_i)^2}{\frac{1}{2}(x_i+y_i)}$ \\
\hline
RBF + Chi-square & $k(x, y) = 1 - \Sigma_{i=1}^n \frac{(x_i-y_i)^2}{\frac{1}{2}(x_i+y_i)} + exp \left(-\frac{\| x-y \|}{2\sigma^2} \right)$ \\
\hline
\end{tabular}
\label{tab:ker_typ}
\end{table}

\vspace{-5mm}
We have used the block-wise coordinate-descent based approach to solve the problem of minimisation given in equation \ref{eq:SML-MFKC} (see \cite{gehler2009feature}, for proof of convexity), as proposed by Xu \etal \cite{xu2013soft}, to obtain the local mimimas, $\vec{\beta}^q$. For  each of the $q$-th kernel, among the $P$ kernels, the best selected feature, $\tilde{V}_q$ is based on the supremum over the set, $\vec{\beta}^q$. The optimal feature-kernel combinations $<M^i, Q^i>$, $\forall i \in \{1,...,P\}$ is thus obtained, where $M^i \in \{\tilde{V}_1, ..., \tilde{V}_F\}$ and $Q^i$ is an element ($k_m$) from the set of $P$ kernels. Each of these features, $ M^i$, are projected in the RKHS using $Q^i$ to obtain a new feature representation to be used in the DA stage.

\subsection{Domain Adaptation}
The proposed DA technique proposed is based on the method described by Hoffman \etal in \cite{hoffman2013efficient}. The normal to the affine hyperplane associated with the k-th binary SVM is denoted as $\theta_k, k = 1, ..., K$, and the offset of that hyperplane from the origin as $b_k$. The authors propose to estimate a transformation $W$ of the input features, or, equivalently, a transformation $W^T$ of the source hyperplane parameters $\theta_k$. Let, $x_1^s, . . . , x^s_{n_S}$ denote the training points in the source domain ($D_S$), with labels $y^s_1, ... , y^s_{n_S}$. Let $x^t_1, . . . , x^t_{n_T}$ denote the labeled points in the target domain ($D_T$), with labels $y^t_1, . . . , y^t_{n_T}$, with hinge loss in this case, denoted as: $\mathcal{L}(y,x,\theta)= \max\{0,1 - \delta(y,k) \cdot x^T \theta\}$. Thus the cost function defined below, can be solved using a coordinate-descent approach.
\vspace{-5mm}
\begin{equation}
\begin{split}
J(W,\theta_k,b_k) & = \frac{1}{2} \|W\|^2_F + \\
 \sum_{k=1}^K \bigg[ \frac{1}{2} \|\theta_k\|^2_2 & + C_S \sum_{i=1}^{n_S} \mathcal{L} \bigg( y_i^s, W \cdot \begin{bmatrix} x^s_i \\ 1 \end{bmatrix}, \begin{bmatrix} \theta_k \\ b_k \end{bmatrix} \bigg) + \\
 & C_T \sum_{i=1}^{n_T} \mathcal{L} \bigg( y_i^t, \begin{bmatrix} x^t_i \\ 1 \end{bmatrix}, \begin{bmatrix} \theta_k \\ b_k \end{bmatrix} \bigg) \bigg] \\
\end{split}
\label{eq:da}
\end{equation}
where, the constant $C_S$ penalizes the source classification error and $C_T$ penalizes the target adaptation error. Minimization of $J$ in equation \ref{eq:da}, yields the transformation matrix $W^T$. The source domain data (features from gallery) are then transformed using $W^T$ to obtain the transformed source domain data.

\begin{figure}[!htbp]
\begin{center}
\includegraphics[scale=0.3]{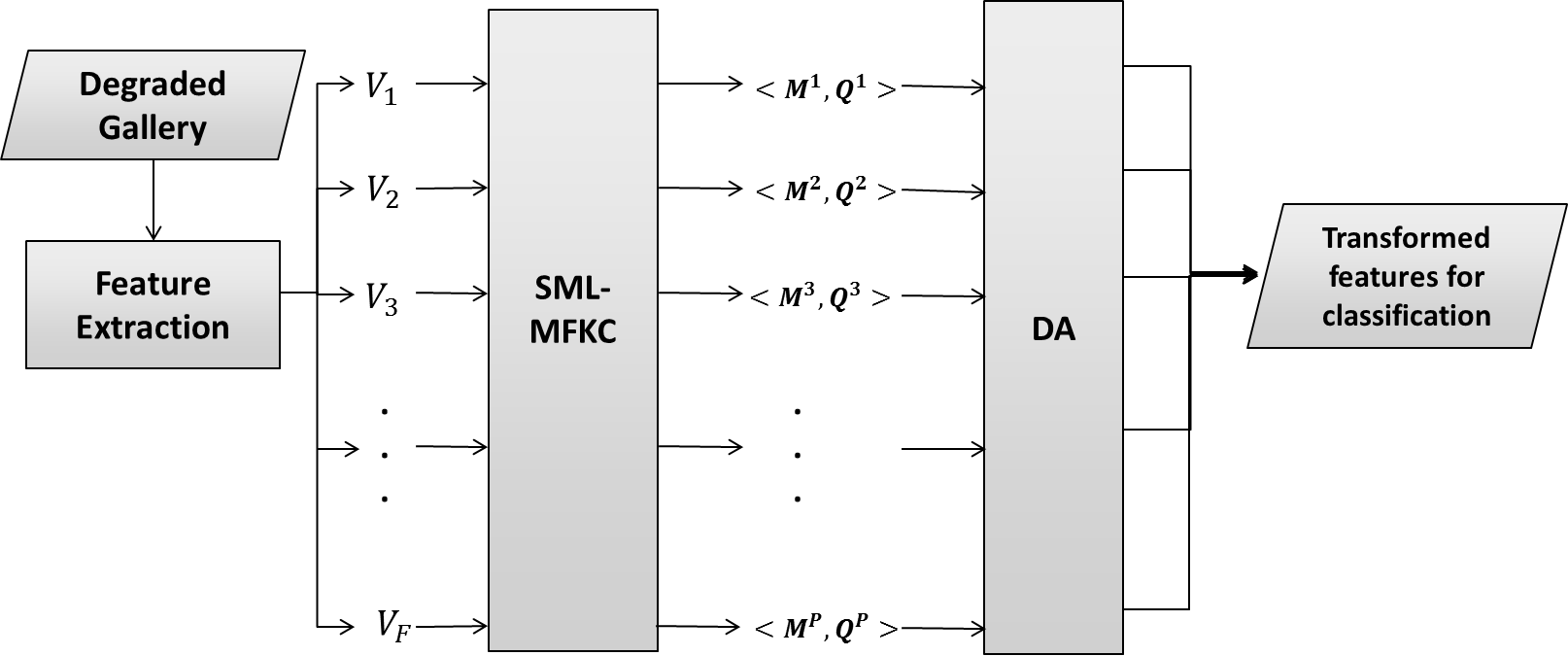}
\end{center}
\caption{Stages in the training phase using SML-MFKC and DA.}
\label{fig:tr}
\end{figure}

The overall training phase for classification is illustrated in figure \ref{fig:tr}, which shows that the features obtained from the degraded gallery samples are fed to the SML-MFKC stage to find the optimal feature-kernel combinations, which in turn is passed into the DA module to obtained transformed source features for classification in RKHS.

\subsection{Classification}
We use the kNN classifier to classify the probe images, which is trained with transformed source domain data based on the transformation obtained from the DA stage. Features from probes samples extracted during the testing stage, are also projected using the respective kernel functions to RKHS, and the Euclidian distance to the nearest transformed source domain data yields the recognized subject ID.

\vspace{-8mm}
\section{Real world surveillance datasets}
\label{sec:data}
For experimentation purpose, we have used three real-world surveillance face datasets, namely, FR\_SURV \cite{rudrani2011face}, SCFace \cite{grgic2011scface} and ChokePoint \cite{wong_cvprw_2011}. For two (SCFace and ChokePoint) out of the three datasets, the probe samples are captured indoor, while the other (FR\_SURV) is shot outdoor. For FR\_SURV dataset \cite{rudrani2011face}, the gallery and probe samples consists cropped face regions at an average of $150 \times 150$ pixels and $33 \times 33$ pixels respectively for $51$ subjects with $20$ samples per class. The SCFace dataset \cite{grgic2011scface} has a huge collection of static images of $130$ different subjects. The gallery cropped by Chehra has an average of $800 \times 600$ pixels resolution, while the cropped probe images at \textit{Distances 1, 2} and \textit{3} has a resolution at an average of $40 \times 40, 60 \times 60$ and $100 \times 100$ pixels respectively, for cams 1-5. Cams 6-8 are not used for evaluation as they are IR images. The ChokePoint dataset \cite{wong_cvprw_2011} consists of $25$ subjects ($19$ males and $6$ females) in portal $1$ and $29$ subjects ($23$ males and $6$ females) in portal $2$. In total, it consists of 48 video sequences and $64,204$ face images. For experimentation, we consider the images obtained from camera, $C1$ as the Gallery set, since it contains maximum frontal images with better lighting conditions than $C2$ and $C3$, which are considered as the probe set. The probe images are passed through all the pre-processing stages, except the face hallucination stage, as the resolution of the images obtained from all these cameras are similar.

\section{Experimental results and discussion}
\label{sec:exp}

Rigorous experimentations have been carried on three real-world datasets; \textit{SCface} \cite{grgic2011scface}, \textit{FR\_SURV} \cite{rudrani2011face}, and \textit{ChokePoint} \cite{wong_cvprw_2011}.  The performance of the proposed methods are compared with several other recent state-of-the-art methods, and the results are reported in table \ref{tab:res}, using Rank-1 Recognition Rate.

\begin{table}[!htbp]
\caption{Rank-1 Recognition Rate for different methods of FR. Results in bold, exhibit the best performance.}
\begin{tabular}{@{}|p{.2cm}|p{2.4cm}|p{1cm}|p{1.4cm}|p{1.6cm}|} 
\hline
% after \\: \hline or \cline{col1-col2} \cline{col3-col4} ...
\textbf{Sl.} & \textbf{Algorithm} & \textbf{SCface \cite{grgic2011scface}} & \textbf{FR\_SURV \cite{rudrani2011face}} & \textbf{ChokePoint \cite{wong_cvprw_2011}}  \\\hline
1 & EDA1 \cite{banerjee2014face} & $47.65$ & $7.82$ & $54.21$ \\\hline
2 & COMP\_DEG \cite{rudrani2011face}& $4.32$ & $43.14$ & $62.59$ \\\hline
3 & MDS \cite{biswas2012multidimensional}& $42.26$ & $12.06$ & $52.13$ \\\hline
4 & KDA1 \cite{banerjee2014face} & $35.04$ & $38.24$ & $56.25$ \\\hline
5 & Gopalan \cite{gopalan} & $2.06$ & $2.06$ & $58.62$ \\\hline
6 & Kliep \cite{kliep} & $37.51$ & $28.79$ & $63.28$ \\\hline
7 & Deep Face \cite{parkhi2015deep} & $41.25$ & $29.35$ & $62.15$ \\\hline
8 & Naive & $77.45$ & $48.23$ & $69.51$ \\\hline
9 & BaseMKL (only VLAD-SIFT) & $53.36$ & $36.54$ & $66.12$ \\\hline
10 & Proposed & $\textbf{79.86}$ & $\textbf{56.44}$ & $\textbf{85.59}$ \\\hline
\end{tabular}
\label{tab:res}
\end{table}

\begin{figure*} [!htbp]
\centering
\includegraphics[scale=0.165]{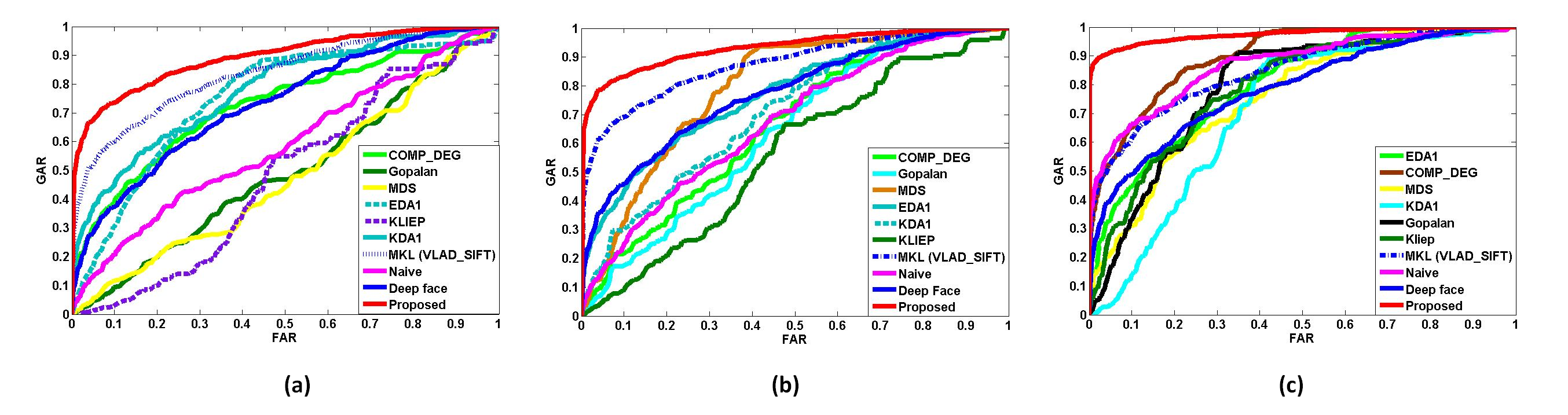}
\caption{ROC plots for performance analysis on: (a) FR\_SURV \cite{rudrani2011face}, (b) SCFace \cite{grgic2011scface} and (c) ChokePoint \cite{wong_cvprw_2011} datasets.}
\label{exp_roc}
\end{figure*}

\begin{figure*} [!htbp]
\centering
\includegraphics[scale=0.165]{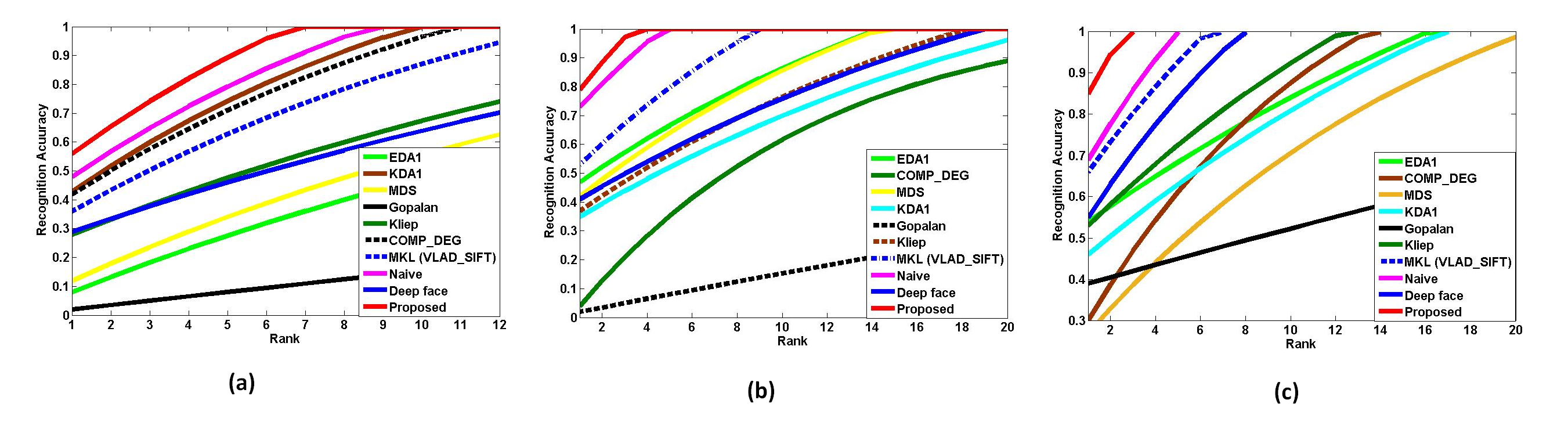}
\caption{CMC plots for performance analysis on: (a) FR\_SURV \cite{rudrani2011face}, (b) SCFace \cite{grgic2011scface} and (c) ChokePoint \cite{wong_cvprw_2011} datasets.}
\label{exp_cmc}
\end{figure*}

Pre-processing is applied uniformly on all face samples prior to application of the algorithms used for comparative study of performance, to obtain fair numbers. For all algorithms in table \ref{tab:res}, code was partially or fully obtained from authors' websites, unless mentioned otherwise. These methods of FR (table \ref{tab:res}), are tuned separately for each of these three datasets, based on our best understanding of the algorithms described in the papers, to make the platform for comparative study uniformly similar (if not identical) across all these methods. In EDA1 method, proposed by Banerjee \etal \cite{banerjee2014face}, DA processing was based on an eigenvector based transformation, whose extension to RKHS is termed as KDA1, in which case the performances on the SCFace \cite{grgic2011scface} and FR\_SURV \cite{rudrani2011face} are comparable; since the non-linearity in the transformation provided by DA technique in RKHS helps to obtain a slightly better result for FR\_SURV compared to that in SCFace.  Rudrani \etal \cite{rudrani2011face} (COMP\_DEG) tries to bridge the gap between the gallery and the probe samples by projecting them both into a lower dimensional subspace determined by the principal components of the feature vectors obtained from the test set (probes). This paper also acts as the primary source for the dataset, FR\_SURV \cite{rudrani2011face}, where an algorithm (over-tuned for degradation in the dataset) is proposed to solve the problem of face recognition under surveillance in an outdoor scenario for specific source and target sensors. Multi-dimensional scaling (MDS) proposed by Biswas \etal in \cite{biswas2012multidimensional} (near implementation to the best of our knowledge) projects both the gallery and the probe samples into a common subspace for classification. The methods proposed by Gopalan \etal \cite{gopalan} and Kliep \cite{kliep} are two DA based techniques used for object classification across domains (codes borrowed from authors). Our proposed method is also compared with the BaseMKL method used in our formulation, but considered here with a single feature representation. This particular (BaseMKL) method, uses VLAD-SIFT \cite{arandjelovic2013all} exclusively as the feature extracted from the face images, and then the optimal kernel is obtained using MKL \cite{bach2004multiple}. We have also experimented on the Deep learning technique proposed by Parkhi \etal \cite{parkhi2015deep}, using the architecture of convolutional neural network (CNN) proposed by the authors. The inputs given to the CNN for training include downgraded gallery images and a few upsampled probes, which were selected as target samples for DA based algorithms. The few set of probes used for DA differ among datasets. Table \ref{tab:sigda} gives the details of the same, used for experimentation. The test (probe) samples includes the total test sets for all cases.  We can observe that our proposed technique has outperformed (our results are given in bold, in Table \ref{tab:res}) all the other competing methods by a considerable margin. The complexity of the datasets is also observed by this result of FR, which are all moderately low (table \ref{tab:res}) in many cases. Minor inconsistencies in performance for competing methods (see rows 2 \& 4, in table \ref{tab:res}) reveal the fact that some algorithms capture the nature (class separability and distributions) of the source and target samples better (due to inherent assumptions) than others, depending on the same for the dataset.

\begin{table} [!htbp]
\caption{Number of samples used as target domain data for DA.}
\begin{tabular}{|c|c|}
    \hline
    \textbf{Dataset} & \textbf{No. of samples used, for DA} \\
    \hline
    FR\_SURV \cite{rudrani2011face} & 5 per subject,\\
    & from 20 random subjects \\
    \hline
    SCFace \cite{grgic2011scface} & 3 per subject,\\
    & from 30 random subjects\\
    \hline
    ChokePoint \cite{wong_cvprw_2011} & 6 per subject,\\
    & from 5 males and 2 females per profile\\
    \hline
\end{tabular}
\label{tab:sigda}
\end{table}

\vspace{-4mm}
Results are also reported using ROC and CMC plots as shown in figures \ref{exp_roc} and \ref{exp_cmc} respectively, for the three datasets used. The plots drawn in red show the performance of our proposed method, which outperform all other competing methods. On an average, the second best performance is given by the naive approach, since the SML-MFKC is also incorporated into it, while the method proposed by Gopalan \etal \cite{gopalan} performs generally the worst. 

A closer look in the table \ref{tab:res} row-wise, reveals that the performance for FR\_SURV dataset produces the least accuracy. This is an indication that there is still further scope of improvement in this field. Also, it shows that this database is quite tough to handle. As we can see that the gallery samples in FR\_SURV are all taken in indoor laboratory conditions, while the probe samples are taken in Outdoor conditions, which results in the large complexity of the database. Since FR\_SURV is an outdoor dataset, we can see the accuracy of FR is less than that of the ChokePoint dataset, which is the easiest to handle among the three. The SCface and the ChokePoint datasets are two indoor (for both gallery and probes) surveillance datasets. Experiments are done in both identification and verification mode. There is still scope of improvement to find a more effective transformation (by DA), such that the distribution of the features of the gallery and the probe become similar. The Naive combination, which  is is very competitive in all the three datasets, consists of only the proposed SML-MKFC process (DA module not used). When these two powerful tools are combined (see figures \ref{fig:fw} \& \ref{fig:tr}), our proposed method outperforms all other methods by an appreciable extent.

\begin{table}[!htbp]
	\caption{Rank-1 Recognition Rate for different methods Deep Learning based FR Methods. Results in bold, exhibit the best performance.}
	\begin{tabular}{@{}|p{.2cm}|p{1.4cm}|p{0.8cm}|p{1cm}|p{1.2cm}|p{1cm}|} 
		\hline
		% after \\: \hline or \cline{col1-col2} \cline{col3-col4} ...
		\textbf{Sl.} & \textbf{Algorithm} & \textbf{\# CNN Layers} & \textbf{FR \_SURV \cite{rudrani2011face}} & \textbf{SCface \cite{grgic2011scface}} & \textbf{Choke Point \cite{wong_cvprw_2011}}  \\\hline
		1 & FV Faces + AlexNet \cite{banerjee2014face} & $8$ & $12.64$ & $35.24$ & $61.59$ \\\hline
		2 & DeepFace \cite{parkhi2015deep}& $19$ & $29.35$ & $41.25$ & $62.15$ \\\hline
		3 & DeepID-2,2+,3 \cite{sun2014deep}& $60$ & $34.94$ & $32.92$ & $66.86$ \\\hline
		4 & FaceNet + Alignment \cite{schroff2015facenet} & $22$ & $36.53$ & $48.21$ & $69.86$ \\\hline
		5 & VGG Face Descriptor + DeepFace \cite{parkhi2015deep} & $19$ & $32.57$ & $46.25$ & $69.86$ \\\hline
		6 & Proposed & $-$ & $\textbf{56.44}$ & $\textbf{79.86}$ & $\textbf{85.59}$ \\\hline
	\end{tabular}
	\label{tab:res_deep}
\end{table}

When compared with the recent state-of-the-art techniques, our method outperforms each of them by a huge margin. The Convolutional Neural Network (CNN) models proposed in the recent past mainly focus on the classification, where the test set and the training set have similar environmental conditions, but are unable to capture the variations in illumination, contrast and scale. Hence, for the three real-world surveillance datasets under experimentation, we can infer that these recent Deep Learning techniques fail to achieve appreciable results as shown in table \ref{tab:res_deep}, compared to our proposed method.

\section{Conclusion}
\label{sec:conc}
The FR under surveillance scenario still remains a open area of research due to its enormous difference in training and the testing conditions. The recent state-of-the-art technique of deep learning  \cite{parkhi2015deep} also fails to perform well for FR under surveillance scenarios, which otherwise have boosted the accuracy of FR in the recent past. An efficient method to tackle the problem of low-contrast and low-resolution in face recognition under surveillance scenario is proposed in this paper, which works well for near-frontal face images only. This paper proposes a novel method using SML-MFKC to obtain an optimal pairing of feature and kernel, followed by DA. The three metrics used to compare the performance of our proposed method with the recent state-of-the-art techniques, show a great deal of superiority of our method than the other techniques, using three real-world surveillance face datasets.

{\small
\bibliographystyle{ieee}
\bibliography{CVPRW_Biom}
}

\end{document}